\documentclass[12pt]{article}
\usepackage[latin1]{inputenc}
\usepackage{graphicx}
\usepackage[english]{babel}
\everymath{\displaystyle}

\title{Mapping weblog communities}
\author{Juan J. Merelo-Guerv\'os  \and Beatriz Prieto \and Fatima Rateb\\
Depto. Arquitectura y Tecnolog\'ia de Computadores,\\
ETS Ingenier\'ia Inform\'atica\\
Universidad de Granada\\
C/ Daniel Saucedo Aranda, s/n\\
  \and
Fernando Tricas\\
Depto. Inform\'atica e Ingenier\'ia de Sistemas\\
C/ Mar\'ia de Luna, 1\\
50018 Zaragoza (Spain) }

\begin{document}

\maketitle

\begin{abstract}
Websites of a particular
class form increasingly complex networks, and new tools are needed to map and understand 
them. A way of visualizing this complex network is by mapping it. A
map highlights which members of the community have similar interests,
and reveals the underlying social network. In this paper, we will
map a network of websites using Kohonen's
self-organizing map (SOM), a neural-net like method generally used for
clustering and visualization of complex data sets. The set of websites
considered has been the
Blogalia weblog hosting site
(based at {\tt http://www.blogalia.com/}), a
thriving community of around 200 members, created in January 2002. In
this paper we show how SOM discovers interesting community features,
its relation with other community-discovering algorithms,
and the way it highlights the set of communities formed over the network.

\end{abstract}

\section{Introduction and state of the art}

Web-based diaries, weblogs (pronounced wee-blogs), or simply
\textit{blogs} \cite{weblogs,blood02,katz99}, have become increasingly popular in the last few
years. Worldwide, there could be several million. Non-English weblogs
are in the hundreds
of thousands\footnote{{\sf http://www.blogcensus.net/} keeps a weblogs
  census; English-language weblogs amount to around one million,
  and the rest of the world, half a million by the time of this
  writing}. Even as weblogs are sometimes perceived as little more 
than post-adolescent rants, they actually are on-screen renderings of
communities of readers/writers, 
which establish long-running relationships; these communities include
weblog owners/writers or editors, people that post comments to weblog
stories, and {\em silent} but persistent readers, both of whom might,
or might not, have its own weblog. A weblog by itself need not
be important, but as part of a community, its importance cannot be
disregarded. All weblogs in the world can be seen as components of a
set of communities, each one with its own idols, axioms, enemies, and
hierarchies. Communities are not clear-cut, since a particular weblog
might belong to several communities at the same time, even though
most weblogs (in fact, all weblogs in the Spanish-speaking community
\cite{blogs2}) are connected to each other by a finite set of links. 

Since blogs perform a sort of collaborative filtering of information
published on the web at large, and are starting to be used as
knowledge management tools, identifying communities becomes specially
important. Information flows more easily within communities than
outside them; getting a message across to as many persons as possible
becomes, then, a matter of identifying communities, and the {\em
 position} of different sites within them. As straightforward as this
view of the  {\em community} concept might seem, the main problem is
that there is no universally accepted definition 
of {\em community} in complex networks. Informally, it can be defined as a set of blogs
(or websites) that share common interests, but this only begs the
definition of {\em common} and {\em interest}. Another possible
definition is to consider a community as a set of blogs that have a stronger
relationship among them than with the rest of the websites of the same
class. Equating {\em relationship} with {\em hyperlinks}
means that a community is a set of weblogs that has more links within
the group than to outside sites. However, while heavily linking
implies belonging to the same community, the inverse does not
necessarily hold: two weblogs\footnote{and its readers/commenters; from now
on, every time we refer to weblogs in a community context, we actually
refer to the group of persons related to that weblog: readers,
writer(s), commenters, and even those that link to it without even
reading it} might both link to the same one, and thus belong, in a
sense, to the same community without being aware of each other or the
community. 

In practice, data available to discover community ascription must be
included in the web page source code, which is text formatted using
HTML tags and some additional meta-tags; sometimes, each text can be
assigned a time-stamp. The 
aforementioned {\em common interest} will have to be identified by
using this data. From the point of view of text content, two
websites are related if they deal approximately with the same
topics. Considering links, two websites are related if
they link to each other in either direction. These two definitions
are actually correlated: Menczer has proved \cite{menczer01} that
pages that link to each other are semantically related. Furthermore, there
are several additional problems with communities related by content:
if a community is defined by keywords, synonyms and hypernims, if not
considered or appropriately chosen, can lead to overseeing certain
websites. This problem is aggravated further by the distinct
characteristics of weblogs as rapidly changing websites and not
focusing on a single topic or set of topics. Using content requires a
vector space representation, usually term frequency/inverse
document frequency \cite{salton83,tfidf}. This representation is
usually highly-dimensional, much 
more so than using links to other members of the set of webs that is
going to be studied. For a small set of sites, link-based
representation is much more compact. Relationship expressed by content
distance, however, is implicit: two weblogs talking about politics,
for instance, need not know each other, although it is very likely
that they do since at least the Spanish blogosphere is connected
\cite{blogs2}. Moreover, in many cases, communities are multilingual;
two weblogs closely related to each other (for instance, written by
the same author) but written in different languages (for instance,
Spanish and Catalan, or Spanish and English) will be completely
unrelated if only content is taken into account.

Meta-content following protocols such as Friend of a Friend (FOAF,
\cite{foaf,dumbill02:foaf}) could, in principle, be also used as
network arcs, but its use is not widespread, and it represents simply
a binary relation (either you are a FOAF or you are not), while links
have some quantitative quality (linking several times is different
from linking only once). 

In this work, links have been chosen over content because they are
easily parseable from the document source; this choice allows for a
low-dimensional representation of each blog which will be represented by
a vector with as many components as blogs in the group under
study. This obviously only holds if the number of relevant sites is
smaller than the vocabulary needed to represent the same sites in a
vector space model. It is
also univocal: a link clearly identifies origin (the weblog it has
been found in) and destination (from the URL). Links represent a real relationship among the
blogs they join: they imply that, at least, one has read the other,
which shows a kind of \textit{community}
relation. This is inferred because communities are created by reading, writing about
other blogs or commenting on them. It is true that there
might be other members of the community not uncovered by this method
(for instance, loyal readers or people who use comments to
participate);  similarly, a member of the community could
be linked to another via a blog not belonging to
the set of blogs under study (Blogalia, in this case); however, we do
not attempt to say the last word about 
community structure in the blogosphere (as is usually called
the set of all weblogs). Our aim is to portray a method to identify
communities by considering hyperlinks a good enough indicator of
community relationship.

Content (distance in vector space) or links (number of
links, or just the existence or not of links) are used to create a complex network
of the set of sites under study; consequently, a community must be
defined by some measure that distinguishes, or makes apart, some sites
from others. There are several possible network structures that could be
considered communities: \textit{cliques}, or sets of sites that link to
each other, \textit{bipartite cliques}, sets of sites which all link to
another, different, set of sites \cite{caldarelli02}, \textit{k-cores
or factions}, sets of sites connected to, at most, \textit{k} other
sites in the group, or \textit{bipartite cores}, which includes both
the connector and the connected sites. Most of these structures can be
computed and displayed with programs such as Pajek\footnote{\label{fnt:ftn4}Pajek
can be downloaded from
{\tt http://vlado.fmf.uni-lj.si/pub/networks/pajek/}} or
UCINET\footnote{UCINET can be downloaded from
{\tt http://www.analytictech.com/}}, but require some initial parameters
such as the number \textit{k} of links or the number of cores we want
to divide the original set into. All
of these are valid definitions, and can be used in some cases. However,
some of them are restrictive in the sense that they only take into account
binary relations, and not the link weight (number of times it has been
used) or direction.
In the case at hand, direction is important: usually, some blog that
has been ``pointed to'' might not even be aware of it\footnote{It is
  very likely that blog authors are aware of incoming links, and there
  are tools, such as {\sf http://technorati.com} or weblog referrer
  logs that allow the author to monitor it}. The
majority of the concepts defined above do not create clear visual
image of the community they are 
describing. 

Sometimes, further steps must be taken to infer complex network
communities. Some of them are geared toward specific communities,
e.g. communities expressed via web pages or email messages, like the
one we are dealing with in this paper. Gibson et
al. \cite{gibson98inferring} proposed one of the first algorithms to
infer web communities; it defined a community as a core of central,
{\em authoritative} pages linked by {\em hub} pages. However, this
definition is a bit fuzzy and does not provide clear-cut partitions of
a set of websites, but it is interesting in the sense that it was one
of the first to realize the importance of communities on the web, and
to propose an algorithm to define them. Shortly afterwards, Flake et
al. \cite{flake00efficient} use a maximum flow/minimal cut algorithm
to define the edges and nodes that act as boundary between
communities.

There exist other algorithms that detect partitions of the original set
according to properties of links, as opposed to properties of
nodes. One of these is the Girvan-Newman algorithm \cite{girvan02},
which detects links that, when removed, 
isolate some part of the original set. Clusters, or communities, are then computed
according to where these removed links are. This 
algorithm discovers communities quite efficiently, as seen in 
\cite{guimera02}, but, once again, it does not discover the
internal structure of each community, or the features that
defines them. 

Recently, Radicchi et al. \cite{radicchi03} review existing community
definition and identification methods, claiming that most community
definitions are algorithm-dependent, and propose a new definition for
community discovery that is independent of the
algorithm. Furthermore, they simplify Girvan-Newman algorithm by using
purely local information to compute edge betweenness. 

This paper, along with our previous work \cite{wbc2004}, uses Kohonen's Self-Organizing Map
\cite{Koh}, which is an unsupervised neural-network like algorithm
that  simultaneously
performs clustering of input data, and maps it to a
two-dimensional surface. Our objective is to demonstrate how the
self-organizing map discovers underlying 
community structure efficiently, allows easy visualization of the
complex network,
highlights the underlying topic that defines each community, and permits
assigning new websites to a community by merely looking at its
links. 

The rest of the paper is organized as follows: first, we make a brief
introduction to Kohonen's self-organizing map in section \ref{bkm:som}. The
next section is devoted to present the results of applying Kohonen's
self-organizing map to community discovery in Blogalia in section
\ref{sec:map}, and, finally, 
our conclusions and an outline of future work is presented.

\section{Kohonen's self-organizing map}
\label{bkm:som}

Kohonen \cite{Koh} originally proposed his self-organizing map
inspired by  
previous work done by von der Malsburg \cite{malsburg73} as a model
for self-organizing 
visual domains in the brain. Kohonen's SOM is composed of a set of
$n$-dimensional vectors, arranged in a 2-dimensional array. Each vector
is surrounded by other 6 (hexagonal) (see figure \ref{seq:refFigure1})
or 8 (rectangular arrangement) 
vectors (see figure \ref{seq:refFigure1}). A size $n$ neighborhood of
a vector is defined as the set of other SOM vectors 
whose index differs in less than a number $n$.

\begin{figure}
\begin{center}
\includegraphics[width=3.6cm,height=2.589cm]{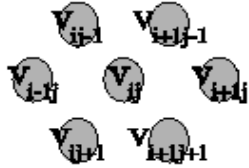}
\caption{\label{seq:refFigure0} Fragment of a self-organizing map, composed
  by 8 {\em neurons} (actually, vectors) neurons, arranged in a hexagonal
neighborhood. Each circle labeled with V represents a vector with the
same dimensions as the input vector in the training set }
\end{center}
\end{figure}

\begin{figure}
\begin{center}
\includegraphics[width=3.874cm,height=2.589cm]{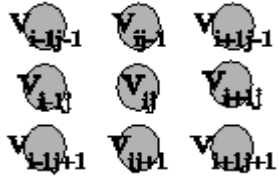}
\caption{\label{seq:refFigure1} Fragment of a self-organizing map with
  square neighborhood.} 
\end{center}
\end{figure}

Kohonen's SOM, as many other heuristic methods, must be
\textit{trained} on the data it is going to model. Training 
proceeds as follow:

\begin{enumerate}
\item A new vector from the training set (the set of data we want to be
modeled) is chosen randomly. 
\item The closest vector, which will be called the {\em winner}, in
  the SOM is computed. 
\item All vectors in the neighborhood of the winner are updated so
that they become closer to the input vector by a factor $\alpha$. 
\item Neighborhood size and $\alpha$ are updated.
\item After a predetermined number of iterations, stop.
\end{enumerate}

The self-organization in the SOMs emerges because different
neighborhoods, not the whole map, are updated every 
time a new vector is 
presented; and the learning
proceeding in an unsupervised way. Other than that, SOM is similar to
any other clustering algorithm such as k-means
\cite{kanungo00analysis}, but, in this case, 
clusters are also arranged geographically. That is why it is said to
perform a topographical mapping. 

Main applications of the self-organizing map are:  

\begin{itemize}
\item \textit{Visualization}: projection from a high-dimensional space to a
two\-dimensional map highlights hidden relationships between data set
members \cite{Bock96}. 

\item {\em Clustering}: unlike other algorithms such as {\em k-means}, each
cluster will be represented by several vectors.

\item {\em Interpolation or function modeling}: it is not
specially suited for this purpose, but if each vector $v$ has
an assigned value $f(v)$, these values can be projected on the
map, and unknown values deduced from it. This is specially useful if
$f(v)$ is actually a vector, or if there might be missing information
from the input set \cite{somcd}. 

\item {\em Classification}: if the original data set is sorted in several
classes, each map vector can be calibrated with a class, and then used
for classification. Even if it is not as efficient for classification
as other neural net algorithms, the fact that it can handle missing
values make it quite useful in those cases. Calibration can be achieved
in several possible ways (using for instance Bayesian criteria), or
additional supervised training using algorithms such as Learning
Vector Quantization (\cite{Kohonen95e}) to improve performance.

\item {\em Vector quantization}: since the map is a model of a data set,
its members can be used to represent that data set, each vector can be
quantized by assigning it to its closest representative in the map.

\end{itemize}

There are many software packages that implement SOM, such as the
\textit{SOM Toolbox }for Matlab, or the \texttt{som} package for R,
but the most popular is probably SOM\_PAK\footnote{\label{fnt:ftn2}The
program is free and can be downloaded from
{\tt http://www.cis.hut.fi/\~{}hynde/lvq/}.}, created originally by
Kohonen's team themselves. This package includes command-line programs
for training and labelling SOMs, and several tools for visualizing it:
\texttt{sammon}, for performing a Sammon projection of data, and
\texttt{umat}, for applying the cluster-discovery UMatrix
  \cite{Ultsch93c} algorithm. We will use these programs in this paper.  

So far, the Kohonen SOM has been used for such diverse applications as
protein secondary structure prediction \cite{jjproteng},
    information retrieval \cite{Kaski97}, rum age visualization
      \cite{rones}, and algorithm visualization \cite{iwann03-gus}. In
      this paper we will take advantage of its capabilities for the
      discovery of communities within Blogalia.

\section{Mapping weblog communities}
\label{sec:map}

The working set of websites corresponds to weblogs
hosted by  Blogalia ({\sf http://www.blogalia.com/}); it hosts around 200 weblogs, of which only
162 actually link or are linked by other weblogs; these are the ones
used in our \ study. All stories, and just the stories (excluding
information in page templates, or dynamic newsfeeds, for instance)
published in Blogalia up to September 
2003 were used for the study; there were around eleven thousand, which
included around seventeen thousand links. Of those, roughly a quarter
were links to other members of the community; this set of links will
be used in this work to try to understand the Blogalia community
structure. Each weblog is
represented by the set of output links to other members of
Blogalia. Of course, and due to this decision, other websites or
weblogs are not considered, which means some sites
closer to some blogs hosted in Blogalia than most of the inhabitants
of that site might be ignored; however, in this paper, our intention
was to discover communities {\em within} Blogalia, not all communities
that included webs hosted by Blogalia.

In this work, each blog is represented by a vector whose
components are the number of times it links to others in Blogalia; if
a blog such as {\tt http://fernand0.blogalia.com/} links to
{\tt http://atalaya.\-blogalia.\-com}/\footnote{\label{fnt:ftn1}Last and
first author's weblogs, respectively} 7 times, the corresponding
element will hold the value 7. Incoming and outgoing links are
considered separately.

UCINET was used to compute {\em factions}, that is, set of blogs
which all point to the same blogs. Results are shown in Table
\ref{tab:fact}. The number of factions was preset to 3. In this case,
the first faction corresponds roughly to the densely connected cluster
shown in figure \ref{seq:refFigure2}; the third, to the sparsely
connected group of blogs, and the second, to all the blogs in
between. 

\begin{table}

\caption{Division of Blogalia into factions, as computed by UCINET.
The number of factions was preset to 3. All the blog URLs are in the
form: {\tt http://NAME.blogalia.com/}, where the name is the string shown
here.\label{tab:fact} }
\begin{center}
\begin{tabular}[c]{|l|p{10cm}|}
\hline
Faction & Components
\\\hline
\#1

&
{\tiny caboclo esbardalladas silly tubo oracle ender pacotilla hazte-escuchar
dragon palabrejas jaio-la-espia dibujante walkyria tse1 saliva mp
bilbao polinesia elforastero superiores terisa simbiosis
ljtarrio yildelen quotidianum gargantua1 oier smith chewie
odisea osito yamato canopus evasivas clio prestige copensar
rimero gargantua peaton aeiou akin eledhwen gnudista paleofreak
jomaweb pawley ciencia15 daurmith jkaranka verbascum blogzine
fbenedetti javarm atalaya www rvr fernand0 }

\\\hline
\#2

&
{\tiny tannhauser cuentacuento qotidianum jarvarm spamzoo russellbeattie
demetro humedadrelativa vendell unhombretranquilo angelina barbara
protoastronomo ocio hunter circulos reval 6cuerdas trunks bontos
fondoazul guetto gripe acuarioland cacharreando electroduende aire
neutrina mayoral miralado ie teo yogurtu amsel xdreus crisei bep
cothinkhealth omar pepino entrelineas sanador exploraciones munchi
borja copensalud planetaneverland confrontacion blojj metro prueba
blogometro }

\\\hline
\#3

&
{\tiny arclnx gofio miatalaya aldor yamisa melicerte latino estilo-005
gaecosita estilo-007 estilo-006 feo riviera kerberos estilo-004 mikel
estilo-05 estilo-001 estilo-003 batiburrillo estilo-002 beta erizoazul
magufos elcubo profes forward isilien maiz elda hispamed cominaii
sieyin kakasico luiso morwen ventanas putten cca pipodols jcohen
cthulhunam rubenlnx robertfernandez mirada escepticismo neuronal
enpelotas hadez desarrollo rivendel hronia }

\\\hline
\end{tabular}
\end{center}
\end{table}

The same data was analyzed using Kohonen's self-organizing map. The
software used was SOM\_PAK version 3.2, with the parameters shown in
table \ref{tab:params}. The algorithm was run 30 times with the same
parameters, but different random initial conditions \footnote{The training set
is available from the authors, with the condition that, if it is used
for any scientific publication, this paper or others by the same
authors, dealing with the same topic, is referenced.}.

\begin{table}
\caption{\label{tab:params}. Parameters used to train Kohonen's
self-organizing map in this paper. The algorithm was run 30 times,
and the map with a minimal squares error was chosen. Values
chosen for these parameters are more or less standard: following
Kohonen's advice, map shape is rectangular, size as small as possible,
and the length of training periods is around 10 and 100 times the size
of the training set.}
\begin{center}
\begin{tabular}[c]{|l|l|}
\hline
{\bf Parameter name} & {\bf Value}

\\\hline
Neighborhood type & Hexa

\\\hline
Neighborhood function & Bubble

\\\hline
Map x size & 9

\\\hline
Map y size & 7

\\\hline
First training period: length & 2000

\\\hline
Neighborhood radius & 9

\\\hline
Training constant & 0.1

\\\hline
Second training period: length & 10000

\\\hline
Neighborhood radius & 1

\\\hline
Training constant & 0.02

\\\hline
\end{tabular}
\end{center}
\end{table}

From the links array, two different analysis were performed: by rows
and columns. Rows represent the set of blogs every blog links to, and
columns represent the set of blogs that links to a particular
one. That means that SOM was applied to blogs represented by
\textit{incoming} and \textit{outgoing } links. On
each map, Umatrix analysis \cite{Ultsch93c}
 was applied: this analysis shows how
the set is clustered, so that \textit{natural} clusters tend
to stand out. 

\begin{figure}
\begin{center}
\includegraphics[width=12cm]{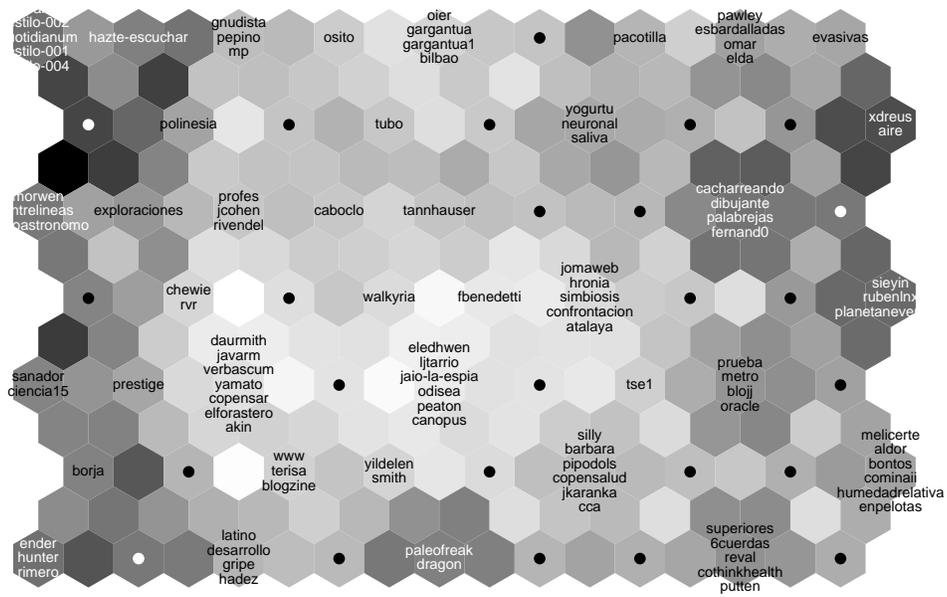}
\caption{ \label{seq:refFigure3}UMatrix map obtained from the SOM
 trained using rows as input, that is, outgoing links. Clusters
 correspond to {\em clear} zones separated by dark hexagons.} 
\end{center}
\end{figure}

\begin{figure}
\begin{center}
\includegraphics[width=12cm]{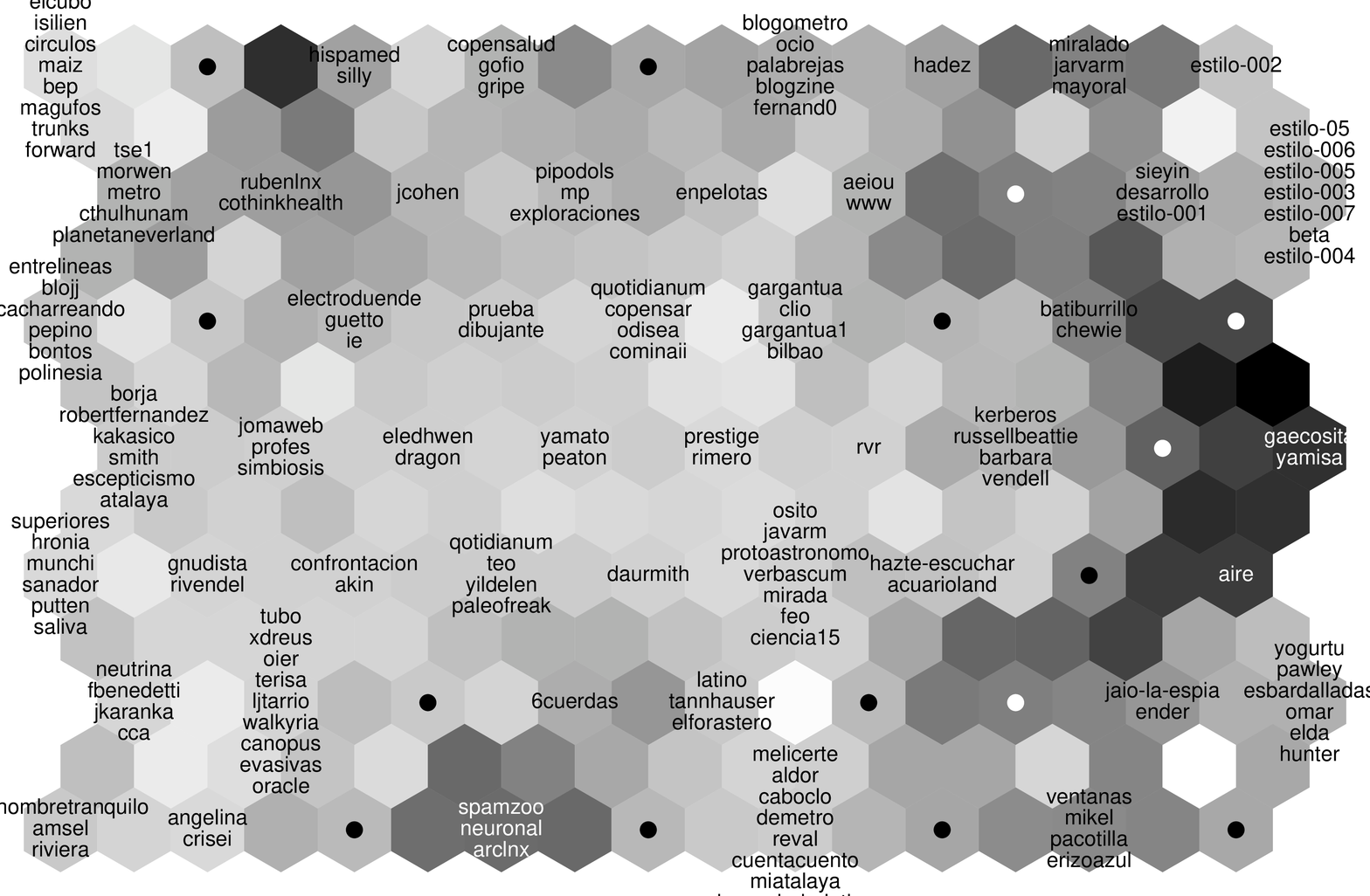}
\caption{\label{seq:refFigure4}UMatrix map obtained form the SOM
trained using columns as input, that is, incoming links.}
\end{center}
\end{figure}

Different results have been obtained by training representing blogs by
incoming or outgoing links. In the first case, shown in figure 5, a
single block, containing the most usually linked-to blogs, stands
out. This block roughly corresponds to the purple core shown in figure
3, and the first faction shown in table 1. The scenario that uses
outgoing links is shown in figure 4 is a bit more discriminating, but,
once again, distinguishes factions and cores as computed by other
methods. 

But it would also be interesting to look at what makes blogs cluster
together in a single node, or what they have in common. It would be
cumbersome to look at each and every node, but, if we look at a couple
of them (for instance, the Southwest corner of figure \ref{seq:refFigure4}, we obtain
the plot shown in figure 6: most of them have a peak of links to
\texttt{pawley}; for instance, \texttt{elda} has a single link, and
it corresponds to that blog; \texttt{pawley} has also many links to
itself, and so on. There are also some other coincidences: a few links
to \texttt{omar}, for instance. 

\begin{figure}
\begin{center}
\includegraphics[width=13cm]{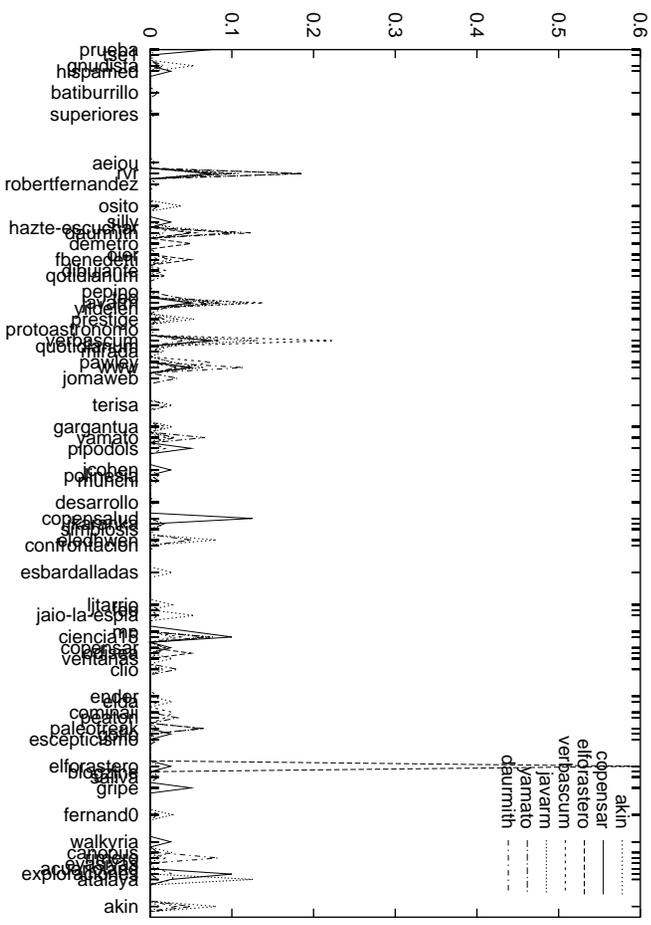}
\caption{ \label{seq:refFigure5} Relative link strength for blogs
  placed in a single node, close to the SE corner, in the incoming
  links map. Sharp peaks, corresponding to a high percentage of links
  to a particular site, are observed {\sf rvr}, {\sf hazte-escuchar}
  and {\sf verbascum}. This node includes also several or the first
  blogs that were created within Blogalia.}
\end{center}
\end{figure}

A similar scenario is seen at the remaining nodes: they have many links to
a blog or set of blogs, which makes the euclidean distance among them
relatively small. That means that the blogs mapped to a single node
roughly correspond to \textit{bipartite cliques} \cite{caldarelli02},
that is, set of nodes whose link pattern is 

To infer communities from this map, a first approach would then be to
assign a community to each node, which would yield several dozens of
communities out of the original hundreds of websites. This is not
satisfactory, however, for two reasons: first, nodes which are closer
in the Kohonen map might also belong to the same community, and
second, some of the blogs that are mapped to a single node do not
actually belong to any community: the upper left corner, for instance,
in figure \ref{seq:refFigure4}
includes all weblogs that do not link to any other.

\begin{figure}
\begin{center}
\includegraphics[width=12cm]{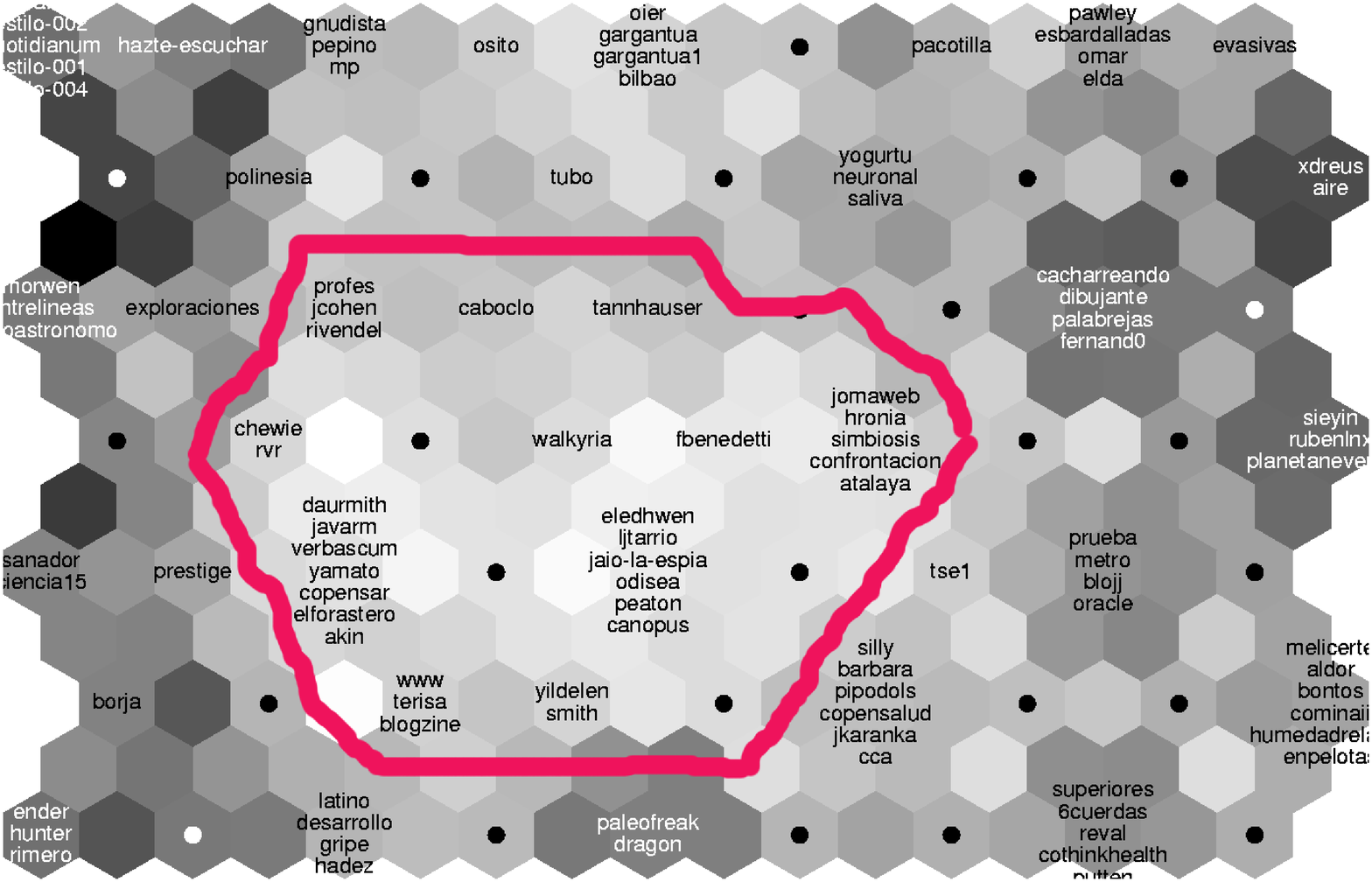}
\caption{ \label{fig:community}UMatrix map obtained from the SOM
 trained using rows as input, that is, outgoing links. Clusters
 correspond to {\em clear} zones separated by dark hexagons. A
 community has been identified and outlined with a red line.} 
\end{center}
\end{figure}

Consequently, we will have to take, a second approach, based on the usual
clustering techniques applied to Kohonen maps postproccessed with the
UMatrix algorithm: clusters are ``white'' zones surrounded by
``black'' boundaries; white zones represent nodes that are close to
each other, while black nodes are far apart from those around it. In
this case, a single community can be appreciated, composed by those
nodes that start roughly with the third row and third column, and end
by the next-to-last row (sixth row) and sixth column. This group of
blogs is outlined in figure \ref{fig:community}.

From this figure, we can gather, in an approach advocated by
\cite{DeBoeck}, that there would be a single cluster, and then smaller
cluster composed by one or, at most, two (the biggest could be one
composed by 4 nodes, right above the outlined cluster). Since this
can be only identified by visual inspection, a new definition of
community cannot be deduced, specially in this case when there is not
a clear-cut division in two or more clusters. So we will introduce a
new definition of community as {\em the set of network nodes that fall
  on the same node of a self-organized map}. This
definition is functional, and, besides, allows assignment of new nodes
just by taking into account its links to the members of the set under
study. An additional advantage is that navigation from a community to
another is possible, just by moving from a node to its neighbors on
the Kohonen map. Besides, a single representative for each community
can be extracted from each node on the network. 

\begin{figure}
\begin{center}
\includegraphics[width=12cm]{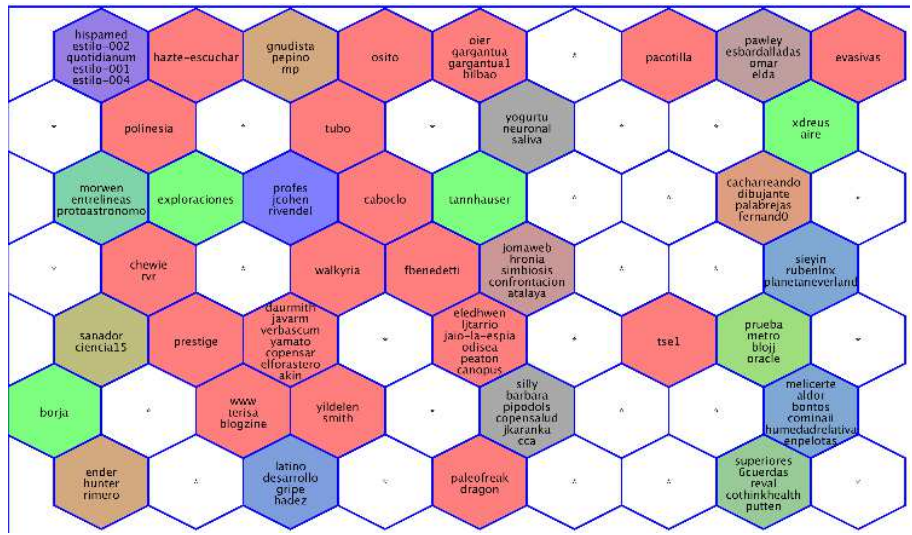}
\caption{ \label{fig:factions} Graphing of factions on the Kohonen
  map trained with outgoing links. ``Red'' faction occupies a large
  part of the map; this red faction corresponds to faction \#1. The
  other factions are not so clearly arranged in the map; this probably
means that they do not really form a community. Green would correspond
to faction \#2, and blue to \#3. Nodes with no blog mapped are left
uncolored.}
\end{center}
\end{figure}

There is indeed some congruence with communities defined this way
and other concepts. In fact, we can represent factions on the Kohonen map, in the
following way: since there are three of them, a primary color (red,
green, blue), will be assigned to each of them; from this, each SOM
node will be assigned an RGB color from the percentage of blogs mapped
to that node belonging to each faction. If blogs belonging to just one
faction are mapped to a node, it will have a primary color; if blogs
belonging to two different factions in equal proportions are mapped to
a node, the color will be 50\%/50\%, for instance, half green, half
red. Results of applying this procedure to the maps are shown in
figure \ref{fig:factions}.

This graph shows that faction \#1 as computed by UCINET is more or less
coherent, and maps in that faction are close to each other, occupying
the majority of the map area. Faction \#1, likewise, forms the core
of this network, being composed mainly by the strongly connected
component of the network; in other words, the strongly connected
component of the network occupies the biggest area in the
self-organizing map. 

More information can be extracted from the self-organizing map. Why is
this layout taken? Why are some blogs in the center, while others
occupy the periphery or corners of the map?

\begin{figure}
\begin{center}
\includegraphics[width=12cm]{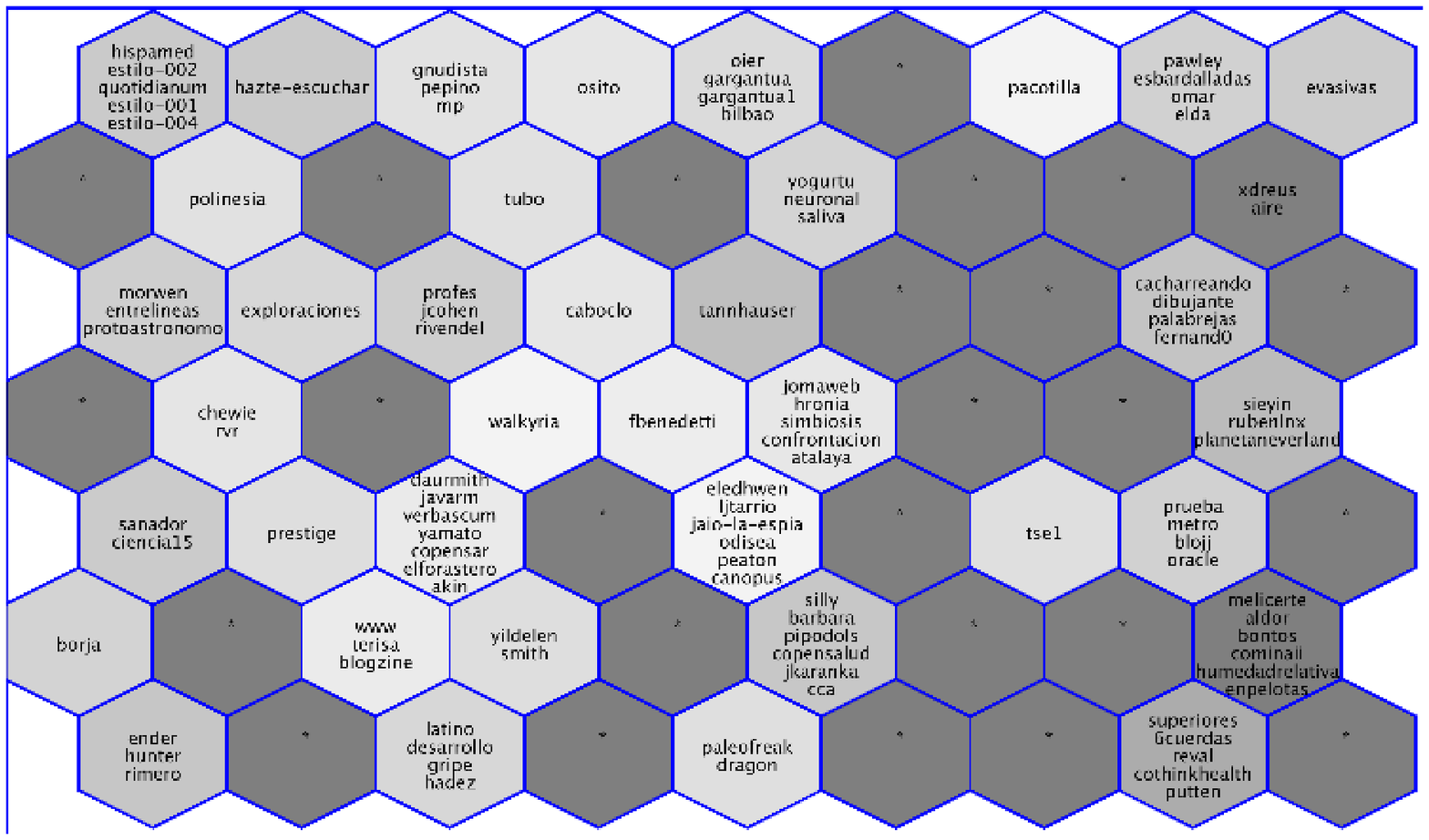}
\caption{ \label{fig:closeness} Graphing of closeness on the Kohonen
  map trained with outgoing links. Gray level corresponds to the
  average closeness of blogs falling on a particular node; the whiter,
  the higher the average closeness is. The node with highest average
  closeness is the one with {\tt eledhwen} and others. Once again,
  nodes with an asterisk do not have any corresponding blog.}
\end{center}
\end{figure}

To answer this, we have plotted average closeness for each node in figure
\ref{fig:closeness}. Apparently, there are some closeness peaks
toward the center of the map, sloping down to the corners, which have
a low average closeness. This is probably the feature that determines
layout, although other measures such as betweenness centrality or
other centrality measures, would have to be investigated.

There is an additional advantage in using Kohonen's self-organizing
map: besides being able to distinguish among different groups, we can
navigate using them. Since we know that blogs mapped to a single node
are close to each other, and are also close to the blogs mapped to the
nodes surrounding them, we could create a path from one blog to
another, or use it as a recommendation for users or writers of a
single blog. Since it works as a mathematical map, another blog, not
belonging to this community, can also be mapped to it just by taking
into account links to the set of blogs already mapped (or links from
them).  

\section{Conclusion}

Web content creation has undergone lately, under the influx of easy
content-management programs such as weblogs, an extraordinary
expansion, which, so far, shows no sign of abating. Interest groups
are created spontaneously among web users, and it is enlightening to
study and identify these groups from the sociological, economical
and technological point of view. Since web-community formation is
generally spontaneous, without an explicit register or inscription by
those that integrate them, and, besides, a particular website might
belong to several communities, one of the first problems posed by its
study is its identification and representation. 

In this paper, we give more details on  using a technique well known
in the pattern recognition and data mining fields: Kohonen's
self-organizing maps; our approach was originally presented in
\cite{wbc2004}. As has been shown in this paper, communities identified by
analyzing self-organizing maps using UMatrix are on a par with those
identified using other techniques, such as \textit{faction} analysis
or \textit{core}  extraction, with the 
additional advantage that \textit{community navigation} can
be achieved by using the map: blogs on the same node, or adjacent nodes,
\textit{belong} (in a fuzzy sense) to the same community. The
self-organizing map, besides highlighting the different communities
and groups present on the sample, make an useful visual
representation. 

The authors of this work intend to continue along one of the following
lines: 

\begin{itemize}
\item
Using self-organizing maps to visualize evolution of a set of blogs,
and the community formation that goes along with it, by mapping
different stages in its life. 

\item Using other algorithms, such as a fuzzy version of Kohonen's
self-organizing map \cite{PascualUM}.  

\item Applying different representations for each blog, using blog content,
instead of blog links: for instance, TFIDF (term frequency/inverse
document frequency) or
latent semantic analysis. 

\item Analysis of nodes with no mapped blog. Do they correspond to
network structural gaps?
  Can they be used to create new blogs that bridge gaps?

\item Analysis of nodes with mapped blogs. What do they represent? 

\item Mapping of complex network measures on the Kohonen map. Can it
  be used to predict any of them, or to offer a fast estimate?
\end{itemize}

\section*{Acknowledgements}

This paper has been funded in part by project TIC2003-09481-C04, of
the Spanish ministry of science and technology, and a project awarded
the Quality and Innovation department of the University of
Granada. Fernando Tricas is with the Group of Discrete Event Systems Engineering
(GISED), and his work is partially supported by TIC2001-1819, financed by the
Spanish Ministerio  de Ciencia  y Tecnologia. We are also grateful to V\'ictor Ruiz for his
creation and continuing support of Blogalia, and his members, for
their support during the realization of this work. 

\bibliography{geneura,NNets-general,wbc,references-Kohonen,CDbib}
\bibliographystyle{unsrt}
\end{document}